\documentclass[twoside,british,3p]{elsarticle}
\usepackage[T1]{fontenc}
\usepackage[utf8]{inputenc}
\usepackage{babel}
\usepackage{orcidlink}
\usepackage{booktabs}
\usepackage{longtable}
\usepackage{amsmath}    % Core math environments and commands (e.g., \text)
\usepackage{amssymb}    % Extra math symbols (e.g., \mathbb, \mathfrak)
\usepackage{amsthm}     % Theorem/Proof environments
\usepackage{algorithm}  % Algorithm environment
\usepackage{algorithmic} % Algorithmic command styles
\usepackage{esvect}     % Fancy vector notation (e.g., \vv)
\usepackage{bbding}     % Symbols like \Checkmark, \XSolidBrush
\usepackage{units}      % Correct spacing for units (e.g., \unit{10}{m})
\usepackage{textcomp}   % Additional text symbols (e.g., \textdegree)
\usepackage{bm}         % Bold math symbols, including Greek letters
\usepackage{listings}
\usepackage{graphicx}   % Graphics inclusion
\DeclareGraphicsExtensions{.pdf,.jpeg,.png,.eps} % file extensions
\usepackage{longtable}  % Tables spanning multiple pages 
\usepackage{booktabs}   % Professional-looking horizontal rules
\usepackage{multirow}   % Spanning cells across multiple rows
\usepackage{array}      % Enhancing array and tabular environments
\usepackage{tabularx}   % Fixed-width table columns
\usepackage{dcolumn}    % Aligning table columns on decimal points
\usepackage{threeparttable} % Notes below a table
\usepackage{pbox}       % Breaks inside table cells
\usepackage{makecell}   % Line breaks and padding in table cells
\usepackage{setspace}   % Used later for \doublespacing
\usepackage{multicol}   % Multi-column text
\usepackage{rotating}   
\usepackage{lscape}     
\usepackage{pdflscape}  
\usepackage[table,xcdraw]{xcolor} 
\usepackage{pagecolor}  % To change the background color of a page
\usepackage{pdfcolmk}   % Related to color marking in PDFs

\usepackage{csquotes}   % Context sensitive (for better quotation marks)
\PassOptionsToPackage{normalem}{ulem}
\usepackage{ulem}       % Underlining/striking out 
\usepackage{blindtext}  % Placeholder text utility
\usepackage{enumitem}   % Customizing lists (itemize, enumerate)
\usepackage{natbib}     % Citation management

\usepackage{hyperref}   % Hyperlinks 
\hypersetup{
    colorlinks=true,
    linkcolor=blue,
    filecolor=magenta,
    urlcolor=cyan,
    pdfborderstyle={/S/U}
}
\bibpunct{(}{)}{;}{a}{,}{,} % Natbib configuration

\usepackage{caption}    % Customizing captions
\usepackage{subcaption} % Sub-figures and sub-tables
\usepackage{tikz}       % Graphics package
\usetikzlibrary{shapes,arrows,positioning,fit,backgrounds,shadows}


\theoremstyle{definition}

\makeatletter
\def\ps@pprintTitle{%
  \let\@oddhead\@empty\let\@evenhead\@empty
  \def\@oddfoot{\footnotesize\itshape Preprint, October 2026\hfill}%
  \let\@evenfoot\@oddfoot}
\makeatother
\begin{document}

\begin{frontmatter}

\title{How Green Is Your Manifesto? Scoring Environmental Positions\\
in the 2024 EU Elections with an LLM Panel}

\author[criep,xaihub]{Franco Corti\corref{cor1}\,\orcidlink{0000-0001-5114-0060}}
\ead{franco.corti@unipd.it}

\author[leeds,xaihub]{Sajid Siraj\,\orcidlink{0000-0002-7962-9930}}
\ead{s.siraj@leeds.ac.uk}

\cortext[cor1]{Corresponding author.}

\affiliation[criep]{organization={CRIEP, Interuniversity Research Centre on Public Economics},
            city={Padova},
            country={Italy}}

\affiliation[leeds]{organization={Leeds University Business School, University of Leeds},
            city={Leeds},
            country={United Kingdom}}

\affiliation[xaihub]{organization={eXplainable Artificial Intelligence (XAI) Hub, University of Leeds},
            city={Leeds},
            country={United Kingdom}}

\begin{abstract}
Locating political parties on environmental and climate policy is central to the study of party competition, yet the established instruments, expert surveys and the hand coding of manifestos, are costly, periodic, and slow to extend to new documents. We ask whether a panel of large language models can serve as a cheaper and reproducible alternative. Treating each manifesto as a set of sentences, three models from independent providers score every sentence on two dimensions, environmental protection and climate change, separating a party's position on an issue from the salience it gives it, and the judgments are averaged to the manifesto level. Agreement across the models is retained as a built-in reliability signal. We apply the panel to 41 manifestos from the 2024 European Parliament election, covering Austria, France, Germany, Italy, Spain, and the European party families, and validate it against the 2024 Chapel Hill Expert Survey for the 25 parties with a like-for-like expert benchmark, using the other manifestos as diagnostic cases. The panel recovers the expert ordering of parties with strong rank agreement on both dimensions, while reading them as systematically greener than experts do, a gap we interpret as the distance between manifesto promise and governing record. Inter-model disagreement modestly flags where the instrument errs, and position and salience prove related for environmental protection but independent for climate change. The result is a transparent, low-cost, and multilingual instrument that complements rather than replaces expert and hand-coded approaches; the prompt, code, and manifesto-level results are released openly.
\end{abstract}


\end{frontmatter}

\section{Introduction\label{sec:intro}}

Environmental and climate policy was one of the defining fault lines of the 2024 European Parliament election. Parties campaigned on the pace of the green transition, the cost of decarbonisation, and the balance between conservation and economic security, and the result reshaped the environmental centre of gravity of the Parliament. Understanding where each party stood is therefore of direct interest to scholars of party competition, to those tracking a possible green backlash, and to anyone modelling the coalitions through which climate legislation must now pass. Yet measuring these positions in a way that is accurate, comparable across countries, and open to scrutiny remains difficult.

The established instruments for locating parties in policy space are expert surveys and hand coding of party documents. Expert surveys such as the Chapel Hill Expert Survey ask country specialists to place parties on policy scales \citep{ches2024}, while the Manifesto Project codes manifestos sentence by sentence against a fixed scheme \citep{manifesto2021}. Both are authoritative and both are costly. They are conducted in periodic waves, depend on scarce expert or coder labour, and are slow to extend to new documents between releases. Expert surveys carry a further subtlety: they record how specialists perceive a party's behaviour, which is not the same thing as what the party's manifesto actually says. For a question about the text itself, an instrument that reads the text directly is preferable.

Large language models make such an instrument conceivable. They can read a manifesto in its original language, reason about individual sentences, and return structured judgments at a fraction of the cost of a survey wave, in a procedure that can be rerun on demand. This promise comes with a well-founded worry. A single model is an unaccountable black box: its scores are sensitive to how the prompt is worded, they can shift silently when the provider updates the model, and one model's idiosyncrasies are invisible from the inside. A measurement that rests on one model offers no internal means of knowing when it should be trusted.

We address this by treating a panel of independent language models as the measurement instrument rather than any single model. Each manifesto is split into sentences, and three models from different providers score every sentence on the same two dimensions using the same prompt, following the ask-and-average approach of \citet{lemens2025}. The per-sentence judgments are pooled into manifesto-level estimates, and the agreement among the models is retained as a native reliability signal: where independent models converge on the same text the estimate is well identified, and where they diverge the estimate is flagged as uncertain. We keep environmental protection and climate change as separate dimensions, and we distinguish a party's position on an issue from the salience it gives it, because a party may speak at length about a policy precisely in order to reject it.

We apply this instrument to 41 manifestos from the 2024 European Parliament election, covering parties in Austria, France, Germany, Italy, and Spain together with the European-level party families, and validate the results against the 2024 wave of the Chapel Hill Expert Survey. Of these, 25 have a like-for-like counterpart in the expert survey and form the validation set; the others are scored with the same instrument and serve as diagnostic cases that expose the practical difficulties of the task. Three questions organise the study. First, can a panel of general-purpose language models, working from the manifesto text, recover expert judgments of party positions? Second, does disagreement among the models flag the cases in which the instrument is least reliable? Third, how does position relate to salience, and can salience be recovered against expert benchmarks at all?

Our answers are broadly encouraging and carefully bounded. The panel recovers the expert ordering of parties with strong rank agreement on both dimensions, while placing parties systematically greener than experts do in absolute terms, a gap we read as the difference between what manifestos promise and what parties are judged to do. Inter-model disagreement carries a modest but real signal about where the instrument errs. Position and salience are related for environmental protection but come apart for climate change, and salience validates only partially against experts. Alongside these findings we contribute a transparent and reproducible pipeline, whose prompt, code, and outputs are released openly, that any team can rerun on new documents. The remainder of the paper reviews the relevant literature (Section~\ref{sec:litrev}), sets out the measurement pipeline (Section~\ref{sec:method}), reports the validation and substantive results (Section~\ref{sec:results}), and discusses the implications and limits of the approach (Sections~\ref{sec:discussion} and~\ref{sec:conclusion}).

\section{Background\label{sec:litrev}}

\subsection{Measuring party positions}

The measurement of party positions in policy space is a long-standing concern of comparative politics, and two instruments dominate. Expert surveys ask country specialists to rate parties on defined scales; the Chapel Hill Expert Survey, now in its 2024 wave, is the standard source for European parties and provides the benchmark we validate against \citep{ches2024}. Hand coding of party documents is the other tradition, exemplified by the Manifesto Project, which assigns every quasi-sentence of a manifesto to a category in a common scheme and aggregates these into positional scales \citep{manifesto2021}. The two approaches have complementary strengths: expert surveys are efficient and draw on deep contextual knowledge, while manifesto coding is grounded directly in what parties commit to in writing. Both, however, are expensive to produce, are released in periodic waves, and cannot easily be extended to a new document without commissioning further expert or coder effort. These constraints motivate the search for methods that scale.

\subsection{Scaling political text automatically}

A substantial literature has sought to estimate positions from text automatically. Early work scaled documents from word frequencies, either by reference to texts of known position, as in Wordscores \citep{laver2003}, or without such anchors, as in the Wordfish model \citep{slapin2008}. Later approaches applied supervised classifiers and, more recently, transformer-based language models to the same task. The current frontier uses large language models as annotators, prompted to label or score text much as a human coder would. Several studies report that such models can approach human coding quality on political and social science tasks \citep{ornstein2025}, and \citet{lemens2025} formalise an ask-and-average protocol in which a model answers many narrow questions about small units of text, with the answers pooled into a stable estimate. Our design sits squarely in this line of work and adopts its sentence-level, ask-and-average logic.

\subsection{Reliability, reproducibility, and the case for a panel}

Using a language model as a measurement instrument raises validity concerns that the political methodology literature is only beginning to address. Model outputs are sensitive to prompt wording, can change when a provider updates a model without notice, and reflect a single system's biases that are not observable from its scores alone. Reproducibility, a basic requirement of a measurement instrument, is not guaranteed when the instrument is a moving and opaque target. Our response is to treat several independent models as a panel rather than to rely on one. This turns the disagreement between models from a nuisance into information: read across models, convergence indicates a well-identified estimate and divergence marks a manifesto on which the reading is uncertain. Pinning each model to a dated version and fixing both the prompt and the input across the panel further ensures that the procedure can be rerun and that any variation in the output reflects genuine differences in judgment rather than differences in what each model was shown.

\subsection{Environmental and climate positions in party competition}

Finally, the substantive object of study shapes the design. Environmental protection and climate change are related but distinct dimensions of party competition, and parties increasingly treat them differently, embracing conservation while resisting climate policy or the reverse. A single environmental score would obscure this, which is why we measure the two separately. The distinction between where a party stands and how much it emphasises the issue is well established in the study of environmental party competition: \citet{carter2013} charts party positions and issue salience across European party systems using expert-survey and manifesto data, and \citet{farstad2018} finds that a party's climate-change salience is driven above all by its left-right ideology, so attention and support are separate quantities that need not move together. Salience deserves the same care: environmental preferences shape vote choice in European Parliament elections, and their electoral weight depends on how salient the issue is \citep{puleo2025}, so we measure salience apart from position rather than folding the two together. It is well established, too, that positions estimated from manifestos diverge systematically from expert surveys, with manifesto-based measures showing a documented centrist bias \citep{ecker2021}; because manifestos are campaign documents written to persuade, they may also read greener than an assessment of what a party does in office, a say-versus-do gap we return to in interpreting our results. Recent work has assembled dedicated corpora for the study of climate discourse in manifestos across many countries \citep{sanford2025}, and it is against this backdrop that we ask how well an automated panel can recover the environmental landscape of the 2024 election.

\section{Methodology\label{sec:method}}

This section describes the measurement pipeline end to end: the corpus, the preparation of the text, the scoring instrument, the panel of language models that applies it, the way individual judgments are aggregated, and the strategy used to validate the results. Two principles run through the design. The first is that every measurement should be traceable to specific sentences in the source text rather than to a model's holistic impression of a document. The second is that the procedure should be replicable, so that an independent team supplying the same manifestos and the same prompt would obtain the same scores.

\subsection{Research design}

We treat each manifesto as a collection of sentences and score every sentence independently on two environmental dimensions. A panel of three large language models applies the same scoring instrument to the same set of sentences, and the per-sentence judgments are averaged into manifesto-level estimates. This follows the ask-and-average approach of \citet{lemens2025}, in which a model is asked many small and well-defined questions rather than one large and ambiguous one, and the answers are pooled.

Three features make the resulting measurement defensible. Scoring at the sentence level keeps each judgment narrow and auditable, because every non-null score is accompanied by the phrase that triggered it. Using a panel rather than a single model exposes and quantifies model-specific idiosyncrasy: agreement across models is evidence of a stable signal, while disagreement is itself recorded as a reliability measure. Fixing both the prompt and the input across models ensures that any variation in the output reflects genuine differences in model judgment rather than differences in what each model was shown.

\subsection{Corpus\label{sec:corpus}}

The corpus consists of 41 manifestos from the 2024 European Parliament election, covering parties in Austria, France, Germany, Italy, and Spain, together with the manifestos of the European-level party families. These five countries were selected because they give broad ideological and geographic coverage of the Parliament while keeping the corpus small enough to be curated and checked by hand. Not every document can be compared with the expert survey, so the corpus is used at two levels. Of the 41 documents, 28 correspond to parties that are also covered by the expert survey. Three of these are multi-party lists without a single expert counterpart (Alleanza Verdi e Sinistra, Besoin d'Europe, and CDU/CSU), while lists led by one surveyed party are matched to that party; this leaves 25 parties as the validation set. The remaining 13 documents, seven European-level manifestos and six national parties outside the survey, are scored with the same instrument but not benchmarked. We retain them deliberately: together with the benchmarked documents, they surfaced most of the practical difficulties reported below, from post-election and very short documents to scanned and bullet-point layouts, and they show how the instrument behaves on texts that an expert survey does not cover. \ref{app:corpus} lists every manifesto with its role in the analysis, the number of sentences scored, and any known issue. Each document is indexed by a controller table that records the party acronym, country, European Parliament group, the identifier used by the expert survey, and the source file, so that every scored manifesto can be linked back to its benchmark entry.

\subsection{Text preparation and sentence segmentation\label{sec:method-text}}

Each manifesto is converted from its published PDF into plain text and then split into sentences. Text is extracted with the PyMuPDF library and segmented with the pysbd sentence boundary detector \citep{pysbd2020}, using its default rules for all languages. Fragments of ten characters or fewer are discarded to remove headings, page furniture, and extraction artefacts, and an upper limit of 3{,}000 sentences per manifesto bounds the cost of very long documents.

The resulting sentence list is produced once per manifesto and shared unchanged across all three models. This is the single most important preprocessing choice in the study. Because the models receive exactly the same sentences in the same order, any dispersion in their scores can be attributed to the models themselves rather than to differences in tokenisation, chunking, or cleaning. A known limitation is that extraction quality depends on the source document: born-digital PDFs segment cleanly, whereas scanned PDFs yield little usable text until they are processed with optical character recognition. One manifesto in the corpus is a scanned document and currently produces far fewer sentences than its length would suggest, an issue we return to in Section~\ref{sec:method-open}.

\subsection{Scoring dimensions and scale}

Each sentence is scored on two dimensions drawn from the expert survey: environmental protection and climate change. Keeping the two dimensions separate is deliberate, because a party can treat them very differently, for example by emphasising conservation or pollution while remaining sceptical of climate policy, and collapsing them into a single measure would hide that distinction.

For each dimension the model reports two quantities. Position, on a scale from 0 to 5, records how supportive the sentence is, ranging from active opposition at one end to placing the issue at the centre of the programme at the other. Salience, on a scale from 0 to 10, records how much the sentence is about the dimension at all. Separating position from salience matters because the two are logically independent: a party may devote many words to an issue precisely in order to reject it, so attention is not the same as support. The model also returns a confidence value, which is recorded as descriptive metadata and not used in aggregation, and, for every non-null score, a short evidence quote in the original language of the manifesto.

A central rule of the instrument is that a null score is not the same as a zero. Null means the sentence is not about the dimension, whereas zero means the sentence engages the dimension and takes an explicitly neutral or opposing stance. This distinction prevents silent sentences from being read as opposition, but it also means that the position scale has no separate neutral midpoint, a feature we return to when interpreting the level of the scores (Section~\ref{sec:results-bias}). How null and zero should subsequently be handled during aggregation is not fully settled, and we return to it in Section~\ref{sec:method-open}.

\subsection{The model panel and prompting protocol\label{sec:method-panel}}

The panel consists of three language models from independent providers, each pinned to a specific version: Claude Sonnet 4.5 (\texttt{claude-sonnet-4-5-20250929}), GPT-4o (\texttt{gpt-4o-2024-11-20}), and Gemini 2.5 Pro (\texttt{gemini-2.5-pro}, with the provider's default reasoning mode). Pinning the versions protects the results against silent model updates and allows the run to be reproduced later. All three are queried at temperature zero, which minimises sampling randomness, so that variation across models reflects disagreement rather than run-to-run noise. Hosted models are not guaranteed to be fully deterministic even at temperature zero, which is one reason the raw outputs of the run are archived (\ref{app:prompt}).

Every model receives the same system prompt, which defines the two dimensions, the two scales, the null rule, and a strict output format. The prompt instructs the model to score each sentence on its own merits, to avoid inferring a position from the party name or its parliamentary group, and to return a single structured object with one record per sentence. This instruction matters, because the measurement should rest on the text rather than on the model's prior associations with a party. In this first iteration, however, each batch of sentences is preceded by a one-line header giving the party acronym, country, expert-survey identifier, and parliamentary group, so the models are told not to use party identity rather than prevented from seeing it; a blinded re-run without the header is the most direct test of how much prior knowledge contributes, and is planned for the next iteration. The full system prompt and the header template are reproduced in \ref{app:prompt}. Sentences are sent in batches of 50 consecutive sentences to fit within the models' context and output limits, and each provider's structured-output mode is used to keep the returned records machine readable. Failed or malformed responses are retried, and long documents are processed in resumable segments so that a run can be interrupted and continued without rescoring completed material.

\subsection{Aggregation and ensemble reliability}

For a given model, the sentence-level scores are aggregated to the manifesto level. A manifesto's position on a dimension is the mean of its non-null position scores, so that only sentences that actually engage the dimension contribute. Its salience is the mean over all sentences, with sentences that do not mention the dimension contributing zero, so that salience reflects the share of the whole document devoted to the issue. This asymmetry between the two measures is intentional, although, as noted above, the precise handling of null values remains an open question.

The three model-level estimates are then combined. The ensemble estimate for a manifesto is the mean of the available model estimates, and the dispersion among the models, measured as their standard deviation, is retained as a reliability signal. We read this dispersion through a multi-criteria lens, treating each model as one criterion or expert judge, so that low dispersion indicates a robust and well-identified position while high dispersion flags a manifesto on which the models genuinely disagree. The number of contributing models is recorded alongside each estimate, because a manifesto scored by fewer than three models carries a weaker estimate and a dispersion that must be interpreted with care.

\subsection{Validation against expert benchmarks}

Manifesto-level estimates are validated against the 2024 wave of the Chapel Hill Expert Survey \citep{ches2024}, a long-running survey in which country specialists place parties on policy scales. The expert scores, which run from 0 (strongly pro-environment) to 10, are rescaled as $(10-x)/2$ to run in the same pro-environment direction and over the same range as the model scores, and the comparison is restricted to the 25 parties of the validation set (Section~\ref{sec:corpus}).

Agreement is assessed with a small set of complementary measures rather than a single statistic, so that different facets of validity remain visible. These include the average signed difference between model and expert scores, which captures systematic bias; the average absolute difference, which captures the typical magnitude of error; the rank correlation between the model and expert orderings, which captures whether parties are placed in the right relative order; the ratio of model dispersion to expert dispersion; and the share of parties whose model and expert scores agree within one standard deviation. Reporting agreement in several ways is deliberate, because a method can reproduce the ordering of parties well while still carrying a constant offset, and only a panel of measures makes that pattern visible.

\subsection{Open methodological questions\label{sec:method-open}}

Several choices in the pipeline are reasonable but not yet fully justified, and we flag them here as directions for further work rather than as settled decisions.

The treatment of null and zero during aggregation is the most consequential. Positions currently ignore null sentences while salience counts them as zero, and although each choice is defensible on its own, the pair has not been tested against alternatives such as imputing neutral values or modelling absence explicitly. A systematic comparison is left to future work.

Model coverage is uneven. One model in the panel applies a noticeably stricter threshold for deciding that a sentence is about a dimension, and therefore returns null far more often than the others. Where it does score, its positions agree with the panel, but its sparser coverage means that the inter-model dispersion for some manifestos partly reflects differences in willingness to score rather than genuine disagreement about a position. Separating these two sources of dispersion is an open problem, and the extreme case, a manifesto that one model scores as entirely off topic, shows why coverage and disagreement should not be conflated.

The aggregation of multi-party lists and coalitions is not yet fixed. Some entries in the corpus correspond to alliances or to European-level families rather than to single national parties, and whether these should be represented by a lead party, by a vote-weighted combination of their members, or excluded from validation is a decision that affects the comparison set and has not been finalised.

Two narrower issues remain. Salience validates weakly against expert judgments by construction, because it is averaged over the whole document and is therefore structurally small, so it is reported alongside position rather than as a primary result. And the reliability of scores from scanned or image-based documents depends on an optical character recognition step that was not applied in this iteration; the documents affected are flagged in \ref{app:corpus}.

\section{Results\label{sec:results}}

The framework produces an environmental protection score and a climate change score for each of the 41 manifestos. The 28 parties that also appear in the expert survey provide the benchmark for validation; excluding the three multi-party lists without a single expert counterpart leaves the 25 parties that form the comparison set reported below. The other manifestos enter the descriptive and reliability analyses but not the validation. We first establish that the ensemble reproduces expert positions, then examine the systematic gap between the two, describe the resulting party landscape, and finally assess the reliability of the ensemble and the relationship between position and salience.

\subsection{Validity against expert positions}

\begin{figure}[htbp]
\centering
\includegraphics[width=\linewidth]{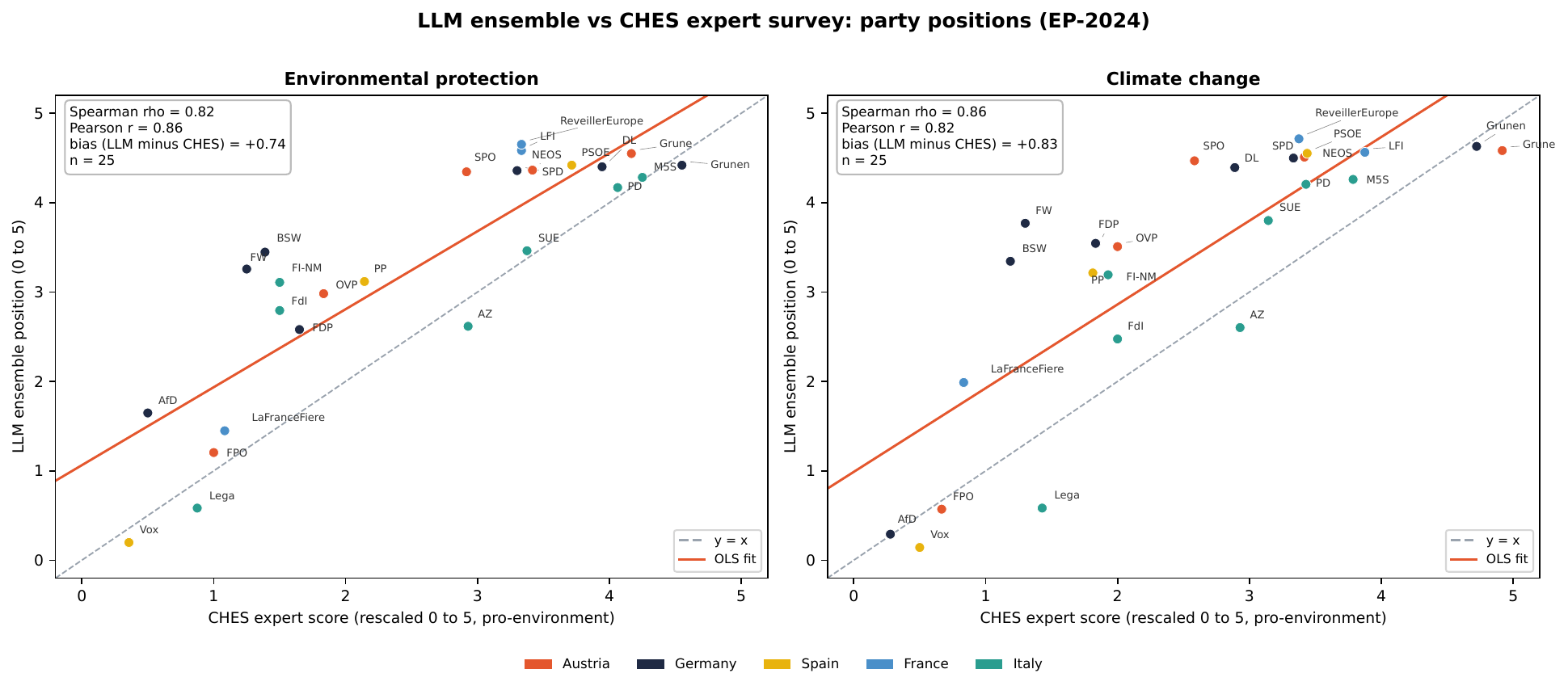}
\caption{Ensemble positions against Chapel Hill Expert Survey scores for environmental protection (left) and climate change (right), for the 25 comparable parties. Points above the dashed line of equality indicate that the ensemble reads the party as greener than the experts do. Colours denote country.}
\label{fig:scatter}
\end{figure}

The ensemble reproduces expert judgments of party positions well on both dimensions. Table~\ref{tab:validation} summarises the comparison against the Chapel Hill Expert Survey, and Figure~\ref{fig:scatter} plots the two sets of scores against each other. The rank correlation between the ensemble and expert orderings is 0.82 for environmental protection and 0.86 for climate change (bootstrap 95\% intervals 0.61--0.92 and 0.68--0.94), and the corresponding linear correlations are 0.86 and 0.82. In practical terms, parties that experts place at the pro-environment end of the scale are placed there by the models as well, and the ordering of parties is recovered almost intact. This answers the study's first question: an ensemble of general-purpose language models, working from the manifesto text, can recover expert positions on environmental and climate policy.

\begin{table}[htbp]
\centering
\caption{Validation of ensemble positions against the Chapel Hill Expert Survey, over the 25 comparable parties. Correlations are between ensemble and expert scores; bias and error are on the shared 0 to 5 scale. The share within one expert standard deviation is computed for the 23 parties for which an expert standard deviation is available.}
\label{tab:validation}
\begin{tabular}{lcc}
\toprule
Metric & Environmental protection & Climate change \\
\midrule
Spearman rank correlation & 0.82 & 0.86 \\
Pearson linear correlation & 0.86 & 0.82 \\
Mean bias (ensemble minus expert) & $+0.74$ & $+0.83$ \\
Mean absolute error & 0.82 & 1.00 \\
Share within one expert standard deviation & 57\% & 39\% \\
Ratio of ensemble to expert dispersion across parties & 1.01 & 1.15 \\
Mean inter-model standard deviation (0 to 5) & 0.29 & 0.35 \\
Parties compared & 25 & 25 \\
\bottomrule
\end{tabular}
\end{table}

\subsection{A systematic pro-environment bias\label{sec:results-bias}}

Although the ordering is recovered, the ensemble does not agree with experts on the absolute level. On average it places parties 0.74 points higher on environmental protection and 0.83 points higher on climate change than the experts do, on the shared 0 to 5 scale. The effect appears in Figure~\ref{fig:scatter} as a consistent shift of the points above the line of equality. Because the shift is systematic rather than random, it inflates the absolute error, which reaches 0.82 points for environmental protection and 1.00 for climate change, while leaving the rank correlation intact.

We read this primarily as a say-versus-do gap. A manifesto is a statement of intent written to appeal to voters, so it naturally reads greener than an expert assessment of how the party actually behaves. Part of the gap may nonetheless come from the instrument itself: the bias varies across the three models reading the same sentences, from $+0.48$ (Claude) to $+0.87$ (GPT-4o and Gemini) on environmental protection and from $+0.51$ to $+1.12$ on climate change, the position scale has no neutral midpoint, and positions are averaged only over sentences that engage the issue. The gap is more pronounced for climate change than for environmental protection, and the share of parties whose ensemble score falls within one expert standard deviation is correspondingly lower for climate, at 39\%, than for environment, at 57\%. The bias is a limitation for any use that needs absolute levels, but it does not undermine the use of the method for ranking or comparison.

\subsection{The party landscape}

The recovered positions follow the expected left to right gradient across all five countries and the European-level lists (Figure~\ref{fig:scatter}). Green, left, and social-democratic parties occupy the pro-environment end, with ensemble positions mostly between 4.2 and 4.7. Christian-democratic parties sit in the middle, around 3.0 to 3.3, while liberal parties are more dispersed, from 2.6 (FDP) to 4.4 (NEOS). Parties of the populist and radical right occupy the low end, with the German AfD at 1.65, the Austrian FP\"O at 1.20, the Italian Lega at 0.58,\footnote{The Lega estimate rests on only 22 sentences because of an OCR layer in the source document (\ref{app:corpus}).} and the Spanish Vox at 0.20.

The two dimensions also come apart in an informative way for some parties. The AfD is the clearest case: its environmental protection position of 1.65 sits well above its climate change position of 0.29, the largest within-party gap in the corpus. This reflects a party that engages with conservation and local environmental language while rejecting climate policy outright. Keeping the two dimensions separate, rather than merging them into a single environmental score, is what makes this pattern visible.

\subsection{Ensemble reliability}

\begin{figure}[htbp]
\centering
\includegraphics[width=\linewidth]{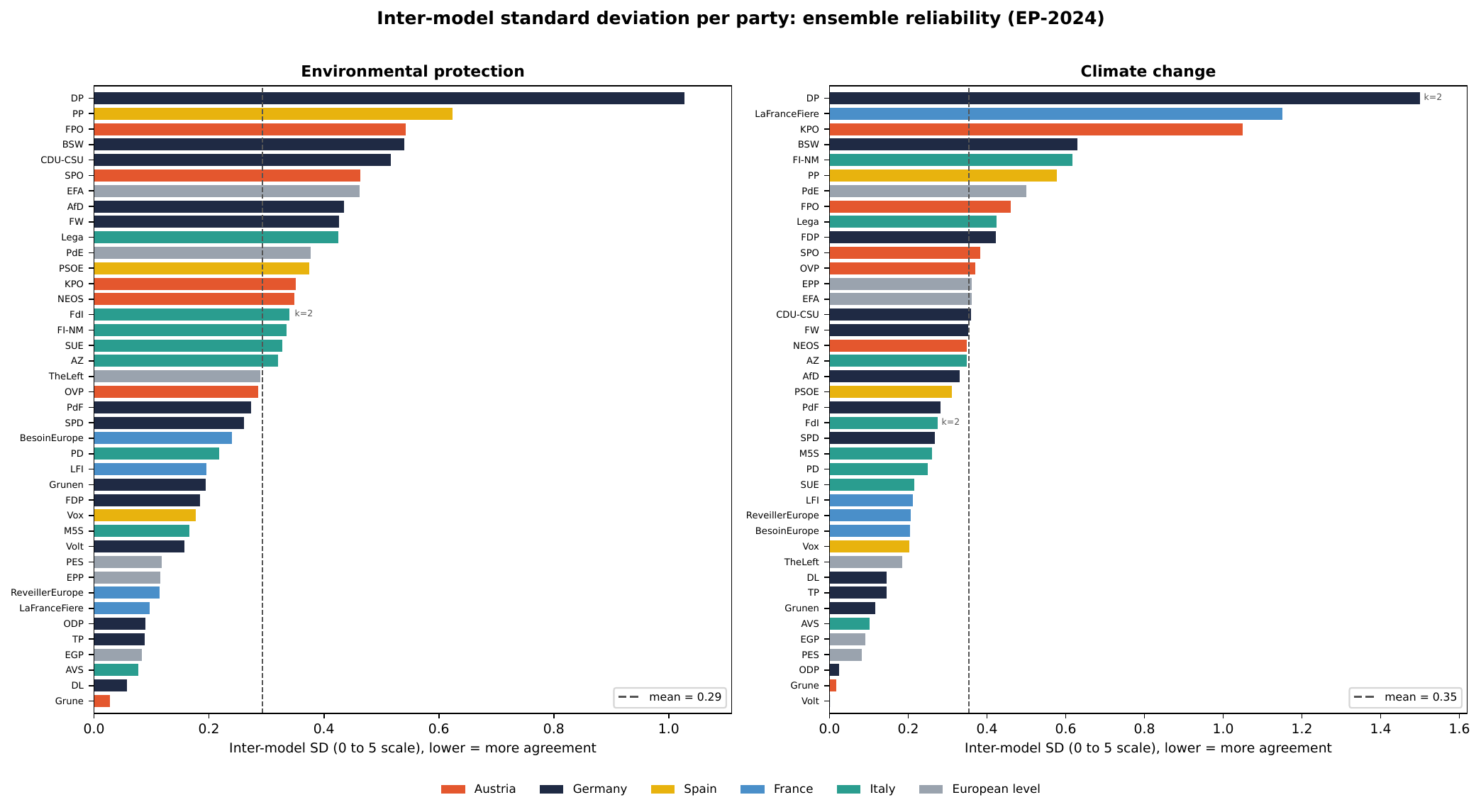}
\caption{Inter-model standard deviation per party for environmental protection (left) and climate change (right), sorted from most to least model disagreement. The dashed line marks the mean. Parties scored by only two of the three models are marked, and their high dispersion partly reflects sparse coverage rather than genuine disagreement.}
\label{fig:sd}
\end{figure}

Agreement among the three models is high. The mean inter-model standard deviation is 0.29 for environmental protection and 0.35 for climate change on the 0 to 5 scale, and for most parties it lies well below these means (Figure~\ref{fig:sd}). Low dispersion indicates that the three models, drawn from independent providers, arrive at similar positions from the same text, which is the reassurance a single model cannot provide.

The largest apparent disagreements are concentrated on a small number of manifestos, and these are exactly the cases in which one or more models scored very few sentences. The German satirical party Die PARTEI (DP), for example, shows the highest dispersion on both dimensions, but each model judged only one or two of its 43 sentences to be on topic, and on climate change only two of the three models returned a score. This points to an important distinction, examined next, between genuine disagreement about a position and an artefact of uneven coverage.

\subsection{Does disagreement flag unreliability?}

If inter-model dispersion is to serve as a confidence signal, it should be larger where the ensemble is further from the expert benchmark. It is, but only moderately. The rank correlation between a party's inter-model standard deviation and its absolute distance from the expert score is 0.36 for environmental protection and 0.43 for climate change (Table~\ref{tab:secondary}). The relationship runs in the expected direction and is statistically clear for climate change, while for environmental protection it falls just short of conventional significance at this sample size. Dispersion is therefore a usable but imperfect indicator: a manifesto on which the models disagree is somewhat more likely to be one on which they are also wrong.

The strength of this signal is held down by the coverage problem noted above. Some of the highest-dispersion cases are driven by a model scoring few sentences rather than by a true conflict of judgment, so part of the dispersion is noise rather than information. The imbalance is large: across the corpus GPT-4o judged only about 4\% of sentences to be on topic (975 of 23{,}932), against 23\% for Claude and 25\% for Gemini, so on the manifestos it barely scored the dispersion reflects its frequent abstention rather than a real conflict of judgment. Separating these two contributions is left to future work, and until it is done the confidence signal should be treated as suggestive rather than decisive.

\begin{table}[htbp]
\centering
\caption{Secondary relationships across parties, reported as Spearman rank correlations. Absolute error is the distance between the ensemble and expert scores. Rows involving the expert survey use the 25 validation parties; the position--salience row uses all 41 analysed manifestos.}
\label{tab:secondary}
\begin{tabular}{lcc}
\toprule
Relationship & Environmental protection & Climate change \\
\midrule
Inter-model standard deviation vs absolute error & 0.36 & 0.43 \\
Position vs salience & 0.45 & 0.11 \\
Ensemble salience vs expert salience & 0.45 & 0.04 \\
\bottomrule
\end{tabular}
\end{table}

\subsection{Position, salience, and structure}

The study's third question concerns how position and salience relate to each other and to the expert benchmark. Within the corpus, the relationship between a party's position and how much attention it devotes to a dimension differs across the two dimensions. The rank correlation between ensemble position and ensemble salience is 0.45 for environmental protection but only 0.11 for climate change (Table~\ref{tab:secondary}). For environmental protection the link is moderate and statistically clear: parties that are more supportive also tend to say more about the issue. For climate change the two are essentially unrelated, so a party hostile to climate policy may still discuss it at length while a supportive party mentions it only briefly. In neither case does salience stand in for position, which is why the two are reported separately.

Salience is also harder to validate against experts than position. The rank correlation between ensemble salience and the corresponding expert salience item is 0.45 for environmental protection, a moderate and statistically clear relationship, but is essentially zero for climate change. Two factors contribute. Salience here is averaged over every sentence in the manifesto and is therefore structurally small, which compresses the differences between parties. In addition, the expert salience items measure a somewhat different notion of issue importance. We therefore treat salience as a secondary output that is recovered only partially, and report it alongside position rather than as a headline result.

\subsection{Robustness to document quality\label{sec:results-robust}}  Several documents in the corpus carry known issues (\ref{app:corpus}), six of them in the validation set. Table~\ref{tab:robust} repeats the validation after excluding these parties step by step. The rank correlations remain between 0.79 and 0.83 for environmental protection and between 0.82 and 0.88 for climate change, and the pro-environment bias persists in every subset. The headline results therefore do not rest on the problematic documents, although the uncertainty widens as the sample shrinks.  

\begin{table}[htbp]
\centering
\small
\caption{Sensitivity of the validation to documents with known issues. Spearman rank correlations with bootstrap 95\% intervals (3{,}000 resamples), and mean bias (ensemble minus expert) on the shared 0 to 5 scale.}
\label{tab:robust}
\begin{tabular}{@{}lccccc@{}}
\toprule
 & & \multicolumn{2}{c}{Environmental protection} & \multicolumn{2}{c}{Climate change} \\
\cmidrule(lr){3-4}\cmidrule(l){5-6}
Parties included & $n$ & Spearman & Bias & Spearman & Bias \\
\midrule
All validation parties & 25 & 0.82 [0.61, 0.92] & $+0.74$ & 0.86 [0.68, 0.94] & $+0.83$ \\
Excluding unverified sources$^{a}$ & 24 & 0.82 [0.60, 0.92] & $+0.69$ & 0.88 [0.71, 0.95] & $+0.76$ \\
\quad and corrupted OCR$^{b}$ & 23 & 0.80 [0.55, 0.92] & $+0.73$ & 0.87 [0.67, 0.95] & $+0.83$ \\
\quad and merged bullet lists$^{c}$ & 21 & 0.83 [0.59, 0.93] & $+0.68$ & 0.86 [0.64, 0.96] & $+0.86$ \\
\quad and truncated or very short$^{d}$ & 19 & 0.79 [0.52, 0.92] & $+0.77$ & 0.82 [0.55, 0.95] & $+0.97$ \\
\bottomrule
\end{tabular}

\smallskip
{\footnotesize $^{a}$Freie W\"ahler. $^{b}$Lega. $^{c}$La France insoumise, Fratelli d'Italia. $^{d}$B\"undnis 90/Die Gr\"unen, Vox.}
\end{table}

\subsection{Summary and limitations}

The ensemble recovers expert positions on both environmental dimensions with strong rank agreement, subject to a consistent pro-environment bias in absolute level. The three models agree closely with one another, and their disagreement carries a modest but real signal about where the ensemble is least reliable. Position and salience are related for environmental protection but independent for climate change, and salience is recovered against experts only partially.

Several limitations qualify these results and motivate the discussion that follows. The validation rests on 25 parties, which is sufficient to establish rank agreement but leaves the secondary correlations with wide margins of uncertainty. The reliability signal is blurred by uneven model coverage, since one model scores far fewer sentences than the others. The treatment of absent versus neutral sentences remains open (Section~\ref{sec:method-open}), the models were shown a party label with each batch of sentences (Section~\ref{sec:method-panel}), and several source documents carry known issues, although excluding them leaves the results essentially unchanged (Section~\ref{sec:results-robust}). Finally, the pro-environment bias, while stable, is not yet independently explained, and separating its measurement and behavioural components is a natural next step.

\section{Discussion\label{sec:discussion}}

The study set out to test whether a panel of general-purpose language models could serve as a measurement instrument for party positions on the environment, and the answer is a qualified yes. Working from manifesto text in its original language, the panel recovers the expert ordering of parties on both environmental protection and climate change with rank correlations above 0.8. This extends the ask-and-average approach of \citet{lemens2025} in two ways that matter for practice: it holds when position and salience are kept as separate dimensions rather than collapsed into one, and it shows that a panel of independent models, not a single model, can be treated as the unit of measurement. For a field that depends on expensive and periodic instruments, an approach that is cheap, multilingual, reproducible, and rerunnable on new documents is a genuine addition to the toolkit, one that complements expert surveys rather than replacing them.

The most telling result is not the agreement but the systematic gap beneath it. The panel reads parties as greener than experts do, by roughly three-quarters of a point on both dimensions, and the shift is consistent rather than random. This is consistent with prior evidence that manifesto-based estimates diverge systematically from expert judgments \citep{ecker2021}, and we read the direction of the gap as the distance between what a manifesto promises and what a party is judged to do: a manifesto is written to win votes, and reads accordingly. The practical lesson is to be precise about what the instrument measures. It recovers relative positions faithfully and is well suited to ranking and comparison, but its absolute levels carry a persuasion premium and should not be read as estimates of behaviour. Whether that premium is stable enough to be calibrated away, and whether it varies across party families, is a question worth pursuing.

The panel design also delivers something a single model cannot: an internal measure of confidence. Where the three models converge the estimate is well identified, and where they diverge it should be treated with caution; we find that disagreement does track error, if only moderately. The honest qualification is that this signal is currently blurred by uneven coverage, because one model scores far fewer sentences than the others and so inflates disagreement for reasons unrelated to any genuine conflict of judgment. Separating true disagreement from coverage artefact is the clearest methodological priority for further work, and resolving it would sharpen the very feature that makes the panel worthwhile.

Finally, keeping position and salience apart proved its value. The two are related for environmental protection but essentially independent for climate change, so a party hostile to climate policy may still discuss it at length, and a single environmental score would have concealed that pattern. Salience nonetheless validates only weakly against expert judgments, partly because it is averaged over the whole document and partly because experts appear to mean something slightly different by issue importance, so we report it as a secondary output. Taken together, these results describe an instrument that is strong where it is asked to rank, honest about where it is uncertain, and careful to keep distinct things distinct.

\section{Conclusion\label{sec:conclusion}}

We have shown that a panel of independent large language models, scoring manifestos sentence by sentence and validated against the Chapel Hill Expert Survey, can recover expert judgments of where parties stood on the environment and climate in the 2024 European Parliament election. The panel reproduces the ordering of parties with strong agreement, exposes a systematic pro-environment gap that we read as the difference between manifesto promise and governing record, and turns disagreement among models into a usable signal of its own reliability. Its limits are equally clear: absolute levels carry a persuasion premium, the reliability signal is blurred by uneven coverage, and salience is only partially recovered. None of these undermines the central claim, that a transparent and reproducible panel of general-purpose models is a credible and low-cost instrument for measuring party positions, complementing expert and hand-coded approaches rather than displacing them. The prompt, code, and results are released openly so that others can apply the pipeline to new elections, new dimensions, and new languages, and the natural next steps are to disentangle coverage from disagreement, to re-run the panel blind to party identity and with improved text extraction, to test whether the persuasion premium can be calibrated, and to widen the corpus beyond five countries.

\section*{Data and code availability}  The scoring prompt and batch header (\ref{app:prompt}), the pipeline code with its run configuration, the manifesto-level results, the corpus metadata with file checksums, and the scripts that reproduce the statistics, figures and corpus table reported here are openly available on Zenodo \citep{corti2026code}, at \url{https://doi.org/10.5281/zenodo.23803120}, under the MIT licence. The Chapel Hill Expert Survey data are available from \url{https://www.chesdata.eu} \citep{ches2024}.

\section*{CRediT authorship contribution statement}

\textbf{Franco Corti:} Conceptualization, Methodology, Software, Data curation, Investigation, Formal analysis, Validation, Visualization, Writing -- original draft, Writing -- review \& editing.

\textbf{Sajid Siraj:} Conceptualization, Methodology, Software, Formal analysis, Validation, Writing -- original draft, Writing -- review \& editing, Supervision.

\section*{Declaration of competing interest}

The authors declare that they have no known competing financial interests or personal relationships that could have appeared to influence the work reported in this paper.

\section*{Declaration of generative AI and AI-assisted technologies in the writing process}

Large language models are the object and the instrument of this study: Claude Sonnet 4.5, GPT-4o, and Gemini 2.5 Pro were used to score the manifesto sentences, as described in Section~\ref{sec:method-panel}. Separately, during the preparation of this work the authors used Claude (Anthropic) to assist with code refactoring and documentation, consistency checks of reported figures and references, and language editing. After using this tool, the authors reviewed and edited the content as needed and take full responsibility for the content of the published article.

\begingroup

\bibliographystyle{elsarticle-harv}
\bibliography{references}
\endgroup

\clearpage 
\appendix 

\setcounter{table}{0}
\section{Corpus\label{app:corpus}}

Table~\ref{tab:corpus} lists the 41 manifestos. The role column distinguishes the 25 parties of the validation set (V), the three multi-party lists matched to the expert survey but excluded from validation (C), and the 13 documents not covered by the survey (X). Sentences are those scored after segmentation and filtering; the on-topic column gives the number of sentences that each model scored on at least one dimension (Claude / GPT-4o / Gemini). Acronyms are those used in the figures.

{\footnotesize
\setlength{\tabcolsep}{4pt}
\begin{longtable}{@{}llp{5.0cm}lcrrc@{}}
\caption{The 41 manifestos of the 2024 European Parliament election in the corpus.}\label{tab:corpus}\\
\toprule
Country & Acronym & Party or list & CHES id & Role & Sentences & On topic & Note \\
\midrule
\endfirsthead
\toprule
Country & Acronym & Party or list & CHES id & Role & Sentences & On topic & Note \\
\midrule
\endhead
\bottomrule
\endfoot
AT & FPO & Freiheitliche Partei \"Osterreichs & 1303 & V & 99 & 18 / 3 / 19 &  \\
AT & Grune & Die Gr\"unen -- Die Gr\"une Alternative & 1304 & V & 1{,}409 & 437 / 63 / 462 &  \\
AT & KPO & Kommunistische Partei \"Osterreichs & -- & X & 200 & 83 / 11 / 77 &  \\
AT & NEOS & NEOS -- Das Neue \"Osterreich & 1306 & V & 614 & 56 / 12 / 66 &  \\
AT & OVP & \"Osterreichische Volkspartei & 1302 & V & 478 & 60 / 6 / 66 &  \\
AT & SPO & Sozialdemokratische Partei \"Osterreichs & 1301 & V & 282 & 48 / 4 / 48 &  \\
DE & AfD & Alternative f\"ur Deutschland & 310 & V & 1{,}215 & 191 / 20 / 215 &  \\
DE & BSW & B\"undnis Sahra Wagenknecht & 313 & V & 467 & 60 / 12 / 68 &  \\
DE & CDU-CSU & CDU/CSU (joint programme) & 301-308 & C & 566 & 58 / 15 / 75 &  \\
DE & DL & Die Linke & 306 & V & 2{,}523 & 627 / 110 / 651 &  \\
DE & DP & Die PARTEI & -- & X & 43 & 1 / 1 / 2 & a \\
DE & FDP & Freie Demokratische Partei & 303 & V & 588 & 70 / 11 / 99 &  \\
DE & FW & Freie W\"ahler & 314 & V & 236 & 33 / 13 / 52 & b \\
DE & Grunen & B\"undnis 90/Die Gr\"unen & 304 & V & 3{,}000 & 717 / 114 / 757 & c \\
DE & ODP & \"Okologisch-Demokratische Partei & -- & X & 705 & 275 / 40 / 294 &  \\
DE & PdF & Partei des Fortschritts & -- & X & 1{,}079 & 131 / 26 / 151 &  \\
DE & SPD & Sozialdemokratische Partei Deutschlands & 302 & V & 997 & 183 / 22 / 207 &  \\
DE & TP & Partei Mensch Umwelt Tierschutz & -- & X & 1{,}442 & 563 / 92 / 697 &  \\
DE & Volt & Volt Deutschland & -- & X & 25 & 3 / 4 / 3 & d \\
ES & PP & Partido Popular & 502 & V & 196 & 16 / 12 / 24 &  \\
ES & PSOE & Partido Socialista Obrero Espa\~nol & 501 & V & 1{,}002 & 227 / 19 / 250 &  \\
ES & Vox & Vox & 527 & V & 32 & 6 / 7 / 7 & a \\
FR & BesoinEurope & Besoin d'Europe (Renaissance and allies) & 626-613-631 & C & 139 & 26 / 9 / 27 &  \\
FR & LaFranceFiere & La France fi\`ere (Reconqu\^ete) & 630 & V & 217 & 33 / 4 / 48 &  \\
FR & LFI & La France insoumise & 627 & V & 188 & 44 / 7 / 40 & e \\
FR & ReveillerEurope & R\'eveiller l'Europe (PS -- Place publique) & 602 & V & 486 & 160 / 21 / 163 &  \\
IT & AVS & Alleanza Verdi e Sinistra & 838-850 & C & 326 & 144 / 34 / 138 &  \\
IT & AZ & Azione -- Siamo Europei & 849 & V & 387 & 142 / 15 / 107 &  \\
IT & FdI & Fratelli d'Italia & 844 & V & 76 & 11 / 0 / 16 & e \\
IT & FI-NM & Forza Italia -- Noi Moderati & 815 & V & 142 & 22 / 16 / 29 &  \\
IT & Lega & Lega Salvini Premier & 811 & V & 22 & 3 / 3 / 4 & f \\
IT & M5S & Movimento 5 Stelle & 845 & V & 1{,}003 & 260 / 77 / 312 &  \\
IT & PD & Partito Democratico & 837 & V & 503 & 75 / 12 / 98 &  \\
IT & SUE & Stati Uniti d'Europa & 832 & V & 121 & 9 / 10 / 8 &  \\
EU & EFA & European Free Alliance & -- & X & 217 & 21 / 23 / 20 &  \\
EU & EGP & European Green Party & -- & X & 1{,}010 & 361 / 50 / 396 &  \\
EU & EPP & European People's Party & -- & X & 624 & 91 / 9 / 98 &  \\
EU & PdE & European Democratic Party & -- & X & 844 & 123 / 35 / 147 &  \\
EU & PES & Party of European Socialists & -- & X & 215 & 37 / 22 / 41 &  \\
EU & PfE & Patriots for Europe & -- & X & 11 & 0 / 1 / 0 & g \\
EU & TheLeft & Party of the European Left & -- & X & 203 & 58 / 10 / 58 &  \\
\end{longtable}
}
\noindent{\footnotesize Notes: (a)~very short document; (b)~regional party publication, to be confirmed as the federal programme; (c)~3{,}347 sentences, of which the first 3{,}000 were scored; (d)~mainly image-based pages; (e)~bullet-point lists without sentence punctuation merged into long units; (f)~corrupted OCR text layer; (g)~document issued after the election. CHES ids joined by a hyphen are the members of a multi-party list.}  

\section{Scoring prompt and technical materials\label{app:prompt}}  All three models received the system prompt below, byte-identical across providers. Each batch of up to 50 consecutive sentences was sent as a user message that begins with a one-line header and an instruction, followed by the numbered sentences:  \lstset{basicstyle=\ttfamily\scriptsize, breaklines=true, columns=fullflexible, frame=single, literate={—}{{---}}1} \begin{lstlisting} PARTY: <acronym> (<country>) | ches_party_id=<id> | ep_group=<group>  Score the following sentences. Return the JSON object only.  1. <sentence> 2. <sentence> ... \end{lstlisting}  

\noindent The system prompt (version 2.0) is reproduced verbatim:  
\begin{lstlisting} 
XAI HUB MANIFESTO SCORING — SYSTEM PROMPT v2.0 (chunked, Python-aggregated)  METHODOLOGY   Framework : XAI Hub Manifesto Prompt System   Method    : Le Mens & Gallego (2025) Ask-and-Average   Unit      : Individual sentences. You score a BATCH of sentences per call.  DIMENSIONS   D4a  Environmental Protection  | Position 0-5 | Salience 0-10   D4b  Climate Change            | Position 0-5 | Salience 0-10    Position 0  = party actively opposes / dismisses environmental/climate protection   Position 5  = party places environmental/climate protection at the centre of its programme   Salience 0  = topic absent from the sentence   Salience 10 = sentence is entirely about this topic  OUTPUT FORMAT — STRICT   Return ONE JSON object and NOTHING else. No prose, no markdown, no code fences.   Schema:   {     "scores": [       {         "sentence_id": <int, matching the number shown>,         "D4a_pos": <0-5 or null>,         "D4a_sal": <0-10 or null>,         "D4b_pos": <0-5 or null>,         "D4b_sal": <0-10 or null>,         "confidence": <1-10>,         "evidence_quote": "<key phrase from the sentence in its original language, or empty string if all scores null>"       }     ]   }    - Do NOT echo back the sentence text. Only the sentence_id and the scores.   - Provide exactly one object per input sentence, preserving sentence_id.   - Do NOT compute any summary. Per-sentence scores only.  RULES   1. Score each sentence independently. Do not adjust for overall party profile.   2. null != 0. null = the sentence is not about that dimension.      0 = the party explicitly takes a neutral or anti-position.   3. evidence_quote required for every non-null score (original language).   4. Do not infer positions from party name or EP group.   5. If ambiguous, lower confidence; never skip a sentence.   6. Output the single JSON object only. \end{lstlisting}  
\noindent The definitive run was carried out in June 2026 with the following settings. 
\begin{itemize}[leftmargin=*,itemsep=1pt] 
\item \emph{Models}: \texttt{claude-sonnet-4-5-20250929}, \texttt{gpt-4o-2024-11-20}, and \texttt{gemini-2.5-pro}, all at temperature zero, with each provider's JSON output mode. 
\item \emph{Text preparation}: extraction with PyMuPDF; segmentation with pysbd using its default rules; fragments of ten characters or fewer discarded; at most 3{,}000 sentences per manifesto. 
\item \emph{Run management}: batches of 50 sentences with per-batch checkpoints; failed calls retried with backoff; malformed JSON recovered sentence by sentence, so that at most an evidence quote is lost. 
\item \emph{Aggregation} (outside the models): position as the mean of non-null scores, salience as the mean over all sentences with null counted as zero, ensemble as the mean of the available models, and dispersion as their standard deviation. 
\end{itemize}  
\noindent The following materials are released on Zenodo with the paper (see the data and code availability statement): the system prompt and batch header above; the scoring, recovery, and consolidation code with its run configuration; the manifesto-level results; the corpus metadata with file checksums; and the scripts that reproduce the validation statistics, figures, and corpus table, including the alignment of the expert-survey scales.

\end{document}